# Analysis of the Displacement of Terrestrial Mobile Robots in Corridors using Paraconsistent Annotated Evidential Logic Eτ

Flavio Amadeu Bernardini[1], Marcia Terra da Silva[1], Jair Minoro Abe[1],
Luiz Antonio de Lima[1] and Kanstantsin Miatluk[2]

[1]Graduate Program in Production Engineering Paulista University,
Sao Paulo, Brazil
[2]Bialystok University of Technology, Bialystok, Poland

## Abstract

*This article proposes an algorithm for a servo motor that controls the movement of an autonomous terrestrial mobile robot using Paraconsistent Logic. The design process of mechatronic systems guided the robot construction phases. The project intends to monitor the robot through its sensors that send positioning signals to the microcontroller. The signals are adjusted by an embedded technology interface maintained in the concepts of Paraconsistent Annotated Logic acting directly on the servo steering motor. The electric signals sent to the servo motor were analyzed, and it indicates that the algorithm paraconsistent can contribute to the increase of precision of movements of servo motors.*

## 1. Introduction

The article focuses mainly on the proposal to develop an algorithm with Paraconsistent Logic for the control of a directional servo motor of an autonomous mobile robot. Given the demand for high investments in new technologies in industrial areas, particularly in the increasing use of robots in the automotive sector [1], this proposal can collaborate with the reduction of the need for maintenance in the factory's internal automated transport.

Regarding mobile robots, they are defined as non-fixed automatic devices capable of moving and interacting with the environment. They are classified according to the environment in which they move, which can be: terrestrial, aerial, aquatic, or underwater [2]. Concerning their displacement environments, they can be in industries [1], where mobile robots can help in the supply of assembly line components; domestic, working in cleaning activities; and at the post office, where they transport items from storage points to distribution points or vice versa.

The design of these robots applies Artificial Intelligence (AI) in decision making. Considering the need for allowing the robot to face uncertainty, alternative systems using non-classical logics, as the so-called Fuzzy systems, are frequently applied. According to [3], paraconsistent logic is one of these non-classical logics, and has applications in software development, neural computing, automation and robotics. In applications that require AI technology in decision-making, the Paraconsistent Artificial Neural Network is already used, including robots [4]. Some published works highlight the pioneering of Paraconsistent Annotated Evidential Logic Eτ applied in a series of robots. The first robot, named Emmy, was built in 1999. After that, in 2004 a second prototype enhanced the first one and a third one was designed in 2009 improving the navigation system. The differential of the present work, when compared to the previous ones, consists in the use of servo motor for the steering control of the prototype of the robot [5].

## 2. BASICS CONCEPTS

The following is a brief approach in the main concepts that based this work.

### 2.1. Servo Motor with Microcontroller

Servo motors are precise, high torque electromechanical devices with a rotating motion proportional to an electrical signal. These movements are monitored by a rotary resistive sensor that has the function of returning the information of the servo's real position to an electronic control circuit [6]. Servo motors can be applied in industrial robotics or precision mechanical machines, such as used in machining centers. The actuator system consists of a direct current motor and gearbox that reduce speed and increase the torque applied to the servo control rod. As it is well known, the main determining factors of the technical characteristics for the correct application of the servo motor are the speed of rotation, the degree of freedom, the torque, the material that makes up the gears, as well as the consumption of electrical energy. Microcontrollers are control devices that can be programmed to meet the requirements of robotics projects using the Integrated Development Environment software.

### 2.2. Design and Control of Mechatronic Systems

The design process of Mechatronic Systems (MS) and other engineering objects, in general, contains several phases [7][8]. In the second phase Conceptual Design (CD) the main design activities aim to generate and evaluate the system's conceptual model, and the main design concept. The Detailed Design (DD) is the third phase and comprehends the mechatronic subsystems' concrete model creation, numeric calculations and, synthesis, and analysis. Lastly, comes the production phase. One of the main requirements for the conceptual model of the MS is that the model should allow the easy transfer from the conceptual description at the CD phase to the concrete models of MS structural and functional design during the DD phase (the synthesis and analysis). MS conceptual model should also take into account MS several levels and present them in a regular formal basis, i.e. lower level – MS structure, current level – MS aggregated dynamic representation as a unit in its environment, higher level – environment construction and technology, MS coordinator and its coordination processes, i.e. design and control.

 Traditional mathematics, AI, and other nowadays models [7][8] do not meet all the above requirements. They do not allow describing robotic and mechatronic systems on all their levels in one common formal basis. So, Hierarchical Systems (HS) technology and created MS model [7-10] are coordinated with known mathematical and AI models, thus meeting all the above requirements. Models of MS structure, MS as a unit in its environment, and MS environment model are presented in the common HS formal basis. The models are connected by HS coordinator, which performs the design and control tasks on its selection, learning, and self-organization strata.

Moreover, conceptual model of MS presents the connected descriptions of MS subsystems of various nature, i.e. mechanical, electrical, and computer. HS technology was implemented in this paper for the case of the Terrestrial Mobile Robot (TMR) conceptual design and control. More attention was paid to the servomotor (electro-mechanical mechatronic subsystem) design and control.

## 2.3. Paraconsistent Annotated Evidential Logic Eτ

Historically, since Aristotelian thought, logic has contributed to correct thinking and in the world observations are not limited to false and true states, and often seeks to relate reasoning with knowledge. Over time, logic has been divided into classical and non-classical, and within the latter, paraconsistent logic has occupied a prominent place, as it deals with the principles of contradiction, in addition to the basic principles of Aristotle's classical logic [11]. Based on the concepts of Paraconsistent Logic, The Paraconsistent Annotated Evidential Logic Eτ works with propositions of type p ($\mu$, $\lambda$), where p is a proposition and $\mu$, $\lambda \in (0, 1)$ (closed range). Intuitively, $\mu$ indicates a degree of favorable evidence and $\lambda$ indicates a contrary degree of evidence of proposition p. Based on the values of the degree of favorable evidence, the degree of unfavorable evidence, the properties of the Paraconsistent Annotated Evidential Logic Eτ are applied to calculate the degree of certainty and degree of uncertainty. Then, these values will be used as a reference for decision making in various applications, such as robotics for example.

## 3. METHODOLOGY

Initially, to ensure a satisfactory sequence in the preparation and execution of this work, the chosen methodology was divided into four stages. The first was a literature search on servomotors, microcontrollers, and Paraconsistent Annotated Logic. Next, conceptual design models of TMR servomotor and its control system were created using HS technology. After these studies, the C language program was prepared for the microcontroller to generate the specific signal to control the servo motor. Tests were performed with the oscilloscope to verify the quality of the signal generated by the microcontroller, as well as to observe the movement of the servo motor. In addition, a logic C programming based on Paraconsistent Annotated Evidential Logic Eτ was applied to verify its efficiency in servomotor decision making. Thus, the utility of logic in controlling the direction of the autonomous robot was verified. In conclusion, the paraconsistent annotated logic in the C programming of the microcontroller was applied to verify the effectiveness of the logic in the decision making of the servomotor, and the results showed good efficiency in controlling the direction of the autonomous robot.

## 4. THE AUTONOMOUS TERRESTRIAL MOBILE ROBOT DESIGN

The robot will be equipped with six ultrasonic sensors connected to a microcontroller that, through a specific electrical signal, will control the robot's directional servo motor.

### 4.1. Mechatronic Design and Control of TMR Servomotor

In this paper, HS technology, and developed MS conceptual model are used for TMR design and control, including all its mechatronic subsystems. In the design process, at the CD phase, the servomotor subsystem was presented by the dynamic system ($\rho,\varphi$) [7, 8], which was transformed to state-space representation at the DD phase, see Figure 1. The final results are presented in the form of equations (1) and (2).

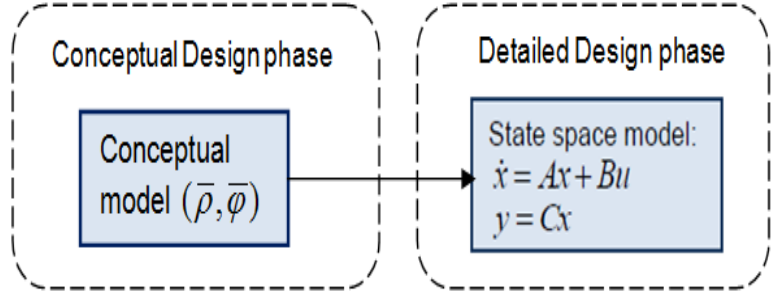

Figure 1. Conceptual model transformation at the design phases.

$$\dot{x} = \begin{bmatrix} \dot{i}_a \\ \dot{\omega}_m \end{bmatrix} = \begin{bmatrix} \dfrac{-L_a}{R_a} & \dfrac{K}{R_a} \\ \dfrac{-K}{B} & \dfrac{-K}{B} \end{bmatrix} \begin{bmatrix} i_a \\ \omega_m \end{bmatrix} + \begin{bmatrix} \dfrac{-1}{R_a} \\ 0 \end{bmatrix} E$$

$$y = \begin{bmatrix} 0 & 1 \end{bmatrix} \begin{bmatrix} i_a \\ \omega_m \end{bmatrix} + 0$$

(1)

The first state equation of (1) corresponds to $\bar{\varphi}$ function, and the output equation corresponds to the reaction $\bar{\rho}$ of $(\bar{\rho}, \bar{\varphi})$ representation at CD phase. State space equations (1) can be transformed at DD phase to the following transfer function (2) if necessary:

$$G(s) = \dfrac{\omega_m(s)}{e_a(s)} = \dfrac{K_i}{s^2 JL_a + sJ_m R_a + K_i K_b} \quad (2)$$

In equations (1) and (2), $i_a$ is armature current, $L_a$ is armature inductance, $R_a$ is armature resistance, $V_a(t)$ is input voltage, $E_b$ is back emf, $K_b$ is voltage constant, $T_L$ is load torque, $T_m$ is motor torque, $\theta_m$, $\omega_m$ are motor angular change and velocity respectively, $K_i$ is a moment constant, $J$ is the motor moment of inertia, $B$ is a friction constant, $K$ is a constant.

The conceptual $(\rho, \varphi)$ model of the servomotor control system was transformed to Paraconsistent Logic model and implemented at DD phase by the developed program unit written in C language. This Paraconsistent Logic program unit was used to control the Dynamixel AX-12A servomotor selected in the design (synthesis) process of TMR. The control results are presented below and show the effectiveness of Paraconsistent Logic model application and the method proposed.

### 4.2. Servo Motor for Robot Control

Therefore, programming the microcontroller in C language can be idealized in the Integral Development Environment after consulting the microcontroller and servo motor manufacturer's manual. The general characteristics of the electrical signal sent by the microcontroller to a servo motor, as well as the respective positions assumed by it, can be seen in Figure 2. In the figure, the first image shows a high signal that lasts 1 ms, followed by a low one. The total period of the signs is the sum of one high and the low that follows. In this case, the configuration of the microcontroller is set to emit signs with a total period of 20 ms, comprising a high-level signal varying its pulse width from 1 ms to 2 ms, and a low-level signal with the corresponding amplitude. To each sign received, the servo motor responds with an angle of movement, varying from 0 to 180 degrees.

In practice, the microcontroller program alternates the high and low levels of the microcontroller output pin to form the signal that will be applied to the servomotor. The high and low-level intervals depend on the load values of the microcontroller time recorder, and the duration of these high-level intervals is controlled using the Positive Duty variable.

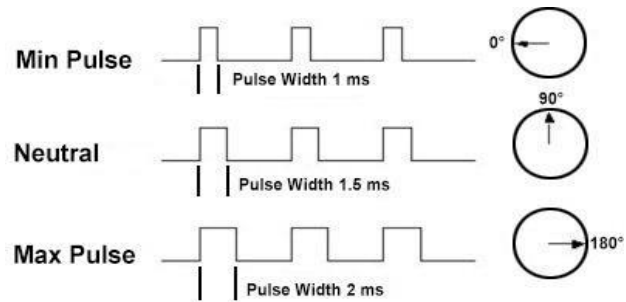

Figure 2. Electric signal control servo motor

## 4.3. Application of Paraconsistent Annotated Logic in Servo Motor to Robots

The initial proposal of the Paranalizer algorithm [12] should be used in the servo motor controller [6] (Dynamixel AX-12A with a set of elements) for the best performance of the angles. The Paranalizer makes subtle adjustments of movement possible, which helps the maintenance of the servo motor working in conformity with the manufacturer's technical guide.

Paranalizer
int paraAnalisador (float mi, float lambda) {
Normalization of the degrees of evidence for the value range between 0 and 1 mi = mi / 100;
Favorable degree of evidence - range of values between 0 and 1 lambda = lambda / 100;
Contrary degree of evidence - range of values between 0 and 1 float Gce = mi - lambda;
Gce - Degree of certainty - Gce = mi - lambda - range of values comprised by - 1 to + 1
float Gin = ((mi + lambda) - 1);
Gin - Degree of uncertainty - Gin = mi + lambda - 1 - range of values comprised by - 1 to + 1
int state = 0;
Extreme and non-extreme logical states - float module_Gce;
Value in the module of the degree of certainty
float module_Gin;
Value in the module of the degree of uncertainty
if (Gce < 0)
module_Gce = Gce * (-1);
else
module_Gce = Gce;
if (Gin < 0)
module_Gin = Gin * (-1);
else
module_Gin = Gin;
Determination of extreme states

Proposition: Free Front
if(Gce >= vcve)
{state = 1};
True - won't hit
else if(Gce <= vcfa)
{ state = 2};
False - will hit - stop, reverse and turn right and then left
else if(Gin >= vcic)
{state = 3};

## 5. RESULTS

The practical tests were satisfactory on microcontroller signals generation to control servomotor, and Figure 3 shows the images captured from the Tektronics oscilloscope model TDS-1002 C-EDU that was used in the tests. The vertical cursors indicate Δt of 2.040ms for the 180º angle and a Δt of 1.020ms for a 90º angle of the servomotor, very close to the values required by the manufacturers.

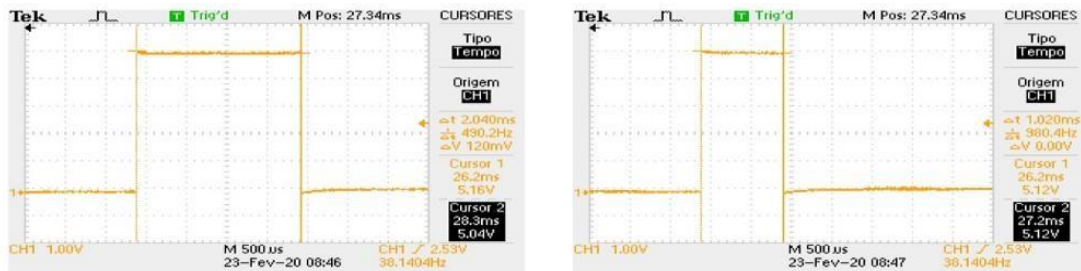

Figure 3. Waveforms generated by the microcontroller for servomotor control.

With the paraconsistent logic applied to the servo motor [8] (Dynamixel AX-12A with a set of elements), it is possible to keep the servo motor within the quality specifications proposed by the manufacturer and help in the displacement (direction) of the land mobile robots in runners. The next program shows the possibility in implementing para- consistent logic in servo control. It can be seen in the examples below that the states of the information can vary depending on the signal of each sensor: True, False, Paracompleteness – information is not sufficient to make a decision, Inconsistent – information is contradictory. This implementation will be complete at the appropriate time of the research. Paraconsistent Logic will be applied when sensors are close to indicate a movement that requires an angle lower then 90º or higher then 180º, as well as long duration of limit angle, which could provoke a possible failure or reduction of the servo motor's life. Examples:

1-Inconsistent - turn slightly to the right, obstacle to the left wide open
else if(Gin <= vcpa)
{state = 4};

2-Paracompleteness - turn slightly left, right obstacle wide open
else if( (Gce >= 0) && (Gce < vcve) && (Gin >= 0) && (Gin < vcic) && (Gce >= Gin))
{state = 5};

Tending
1-Almost true tending to inconsistent - turning too much to the right, obstacle to the left next -

turning more than state 3
else if((Gce >= 0) && (Gce < vcve) && (Gin >= 0) && (Gin < vcic) && (Gce < Gin))
{state = 6};

2-Inconsistent tending to true - turn slightly right, obstacle left open - turn less than state 5
else if((Gce >= 0) && (Gce < vcve) && (Gin > vcpa) && (Gin <= 0) && (Gce >= modulo_Gin))
{state = 7};

3-Almost true tending to Paracompleteness - turning a lot to the left, obstacle to the right next - turning more than the state 8
else if((Gce >= 0) && (Gce < vcve) && (Gin > vcpa) && (Gin <= 0) && (Gce < modulo_Gin))
{state = 8};

4-Paracompleteness tending to the true - turn left slightly, obstacle open right - turn more than state 4
else if((Gce > vcfa) && (Gce <= 0) && (Gin > vcpa) && (Gin <= 0) && (modulo_Gce >= modulo_Gin))
{state = 9};

5-Almost false tending to paraconsistent - stop turning too much to the left - almost hitting, an obstacle to the right too close
else if((Gce > vcfa ) && (Gce <= 0) && (Gin > vcpa) && (Gce < Gin) && (Gin <= 0))
{state = 10};

6-Paracompleteness tending to false - stop and turn a little to the left, obstacle to the right open very close
else if((Gce > vcfa) && (Gce <= 0) && (Gin >= 0) && ( Gin < vcic) && (Gce >= Gin))
{state = 11};

7-Almost false tending to inconsistent - stop and turn too much to the right, an obstacle to the left too close
else if((Gce <= 0) && (Gce < vcfa) && (Gin >= 0) && (Gin < vcic) && (Gce < Gin))
{state = 12};

8-Inconsistent tending to false - stop and turn slightly to the right, an obstacle to the left open too close
{return state};

## 6. CONCLUSION

The work showed the applicability of the developed algorithms based on Paraconsistent Annotated Evidential Logic Eτ algorithms in the DD phase in the robot's servomotor and, thus, contributes to the servomotor's efficiency and assist in driving decision making. Additionally, the application of Paraconsistent Logic allows to maintain the servomotor working within the manufacturer specifications, which contributes for a longer life cycle. As future work, the algorithm must be improved to ensure the use of the servomotor within its technical specifications and keep the perspective of the device's life. This article has provided possibilities that will be explored in the next phase with new experiments extensively.

ACKNOWLEDGEMENTS

"This study was financed in part by the Coordenação de Aperfeiçoamento de Pessoal de Nível Superior - Brasil (CAPES) -Finance Code 001"

AUTHORS

**Flavio Amadeu Bernardini** holds a degree in Mathematics - Integrated College of Science as Human, Health and Education of Guarulhos (2007). Specialization course in Industrial Automation - National Industry Service (SENAI) College of Tecnologia Mechatronics (2015). He is currently an instructor of professional practices in the area of Electronics - SENAI - Regional Department of Sao Paulo. He has experience in electronic maintenance. Studying with a Master's program in Production Engineering - Paulista University – UNIP.

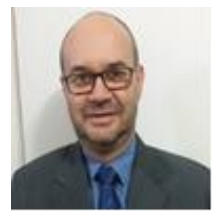

**Márcia Terra da Silva** is Full Professor at the Post Graduation Program of Production Engineering of Universidade Paulista (PPGEP-UNIP). In the last 20 years, she has been developing research in the area of Service Operations Management, with focus on professional services as healthcare and educational services, and has published several articles in Brazilian and international journals. Currently, she researches High Education organization and management, with leading interest in workers' qualification to the demands of the Industry 4.0.

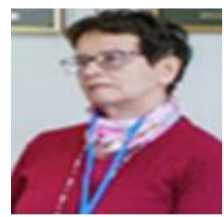

**Jair Minoro Abe** received B.A. and MSc in Pure Mathematics - University of Sao Paulo, Brazil. Also received the Doctor Degree and Free-Teacher title from the same University. He is currently coordinator of Logic Area of Institute of Advanced Studies - University of Sao Paulo Brazil and Full Professor at Paulista University - Brazil. His research interest topics include Paraconsistent Annotated Logics and AI, ANN in Biomedicine and Automation, among others. He is Senior Member of IEEE. 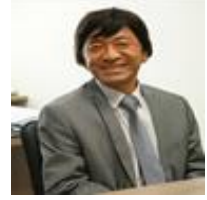

**Luiz Antonio de Lima** is Doctor of Science student in Production Engineering Paulista University, Master degree in Production Engineering in the area of Artificial Intelligence Applied to Software Paraconsistent Measurement Software, Post-Undergraduate Degree in EAD, University Professor, General Coordinator of IT Course and Campus Assistant: (2008-2009). University Professor: 02 Postgraduate Course and 12 Higher Courses in 43 disciplines; Speaker and Event Organizer: SENAED; NETLOG; Wics; WINFORMA. 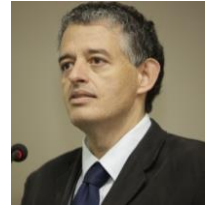

**Kanstantsin Miatliuk** is a Professor of Bialystok University of Technology, Poland. He received his M.Sc. degree in Robotics from the Belarusian State Technical University, Minsk, Belarus (1988), his Ph.D. degree in Automation and Robotics from AGH University of Science and Technology, Krakow, Poland (2006) and his D.Sc. degree in Machines Constructing in Warsaw, Poland (2019). He is IEEE member. K.Miatliuk was a visiting professor in Kyung Hee University, Korea (2008, 2010) University of Southern Denmark, Denmark (2012) and University of Las Palmas Gran Canaria, Spain (2016). K.Miatliuk participated in numerous EU R&D projects. His research interests include mechatronics and robotics, systems science and computer science. 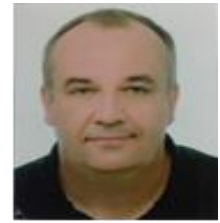